\documentclass[10pt, conference, compsocconf]{IEEEtran}
\IEEEoverridecommandlockouts

\usepackage[utf8]{inputenc}
\usepackage[english]{babel}
\usepackage{graphicx}
\usepackage{xcolor}
\usepackage{cite}
\usepackage[T1]{fontenc} 
\usepackage{amsmath}
\usepackage{multirow}
\usepackage[linesnumbered,ruled]{algorithm2e}

\usepackage{bm}
\usepackage{array}
\usepackage{url}
\usepackage{psfrag}
\usepackage{mathtools}
\usepackage{soulutf8}
\usepackage{amssymb}

\SetKwInput{KwInput}{Input}                
\SetKwInput{KwOutput}{Output}              
\SetKwInput{kwInit}{Initialize}

\usepackage{caption}

\usepackage{soul}
\usepackage[linesnumbered,ruled]{algorithm2e}
\usepackage{titlesec}
\usepackage[acronym]{glossaries}
\usepackage{comment}
\titlespacing{\section}{0pt}{\parskip}{-\parskip}

\newacronym{ai}{AI}{artificial intelligence}
\newacronym{bs}{BS}{base station}
\newacronym{agv}{AGV}{autonomous guided vehicle}
\newacronym{ml}{ML}{machine learning}
\newacronym{pid}{PID}{propotional-integral-derivative}
\newacronym{px}{px}{pixels}
\newacronym{fov}{FoV}{field-of-view}

\newacronym{roi}{RoI}{region-of-interest}

\newacronym{trl}{TRL}{technology readiness level}

\begin{document}

\title{Codesign of Edge Intelligence and Automated Guided Vehicle Control
}
\author{
\IEEEauthorblockN{
Malith~Gallage,~\IEEEmembership{Student Member,~IEEE,}, 
Rafaela~Scaciota,~\IEEEmembership{Member,~IEEE,},
Sumudu~Samarakoon,~\IEEEmembership{Member,~IEEE},
\\and
Mehdi Bennis~\IEEEmembership{Fellow,~IEEE}
}
\IEEEauthorblockA{
	\small%
	Centre for Wireless Communication, University of Oulu, Finland \\
	email: \{malith.gallage,rafaela.scaciotatimoesdasilva,sumudu.samarakoon,mehdi.bennis\}@oulu.fi 
}
\vspace{-20pt}
\thanks{
The authors would like to acknowledge the support and contributions of Abdulmomen Ghalka, Dinesh Manimel Wadu, Mithila Amarasena, and Suranga Prasad for developing the image processing solutions. 

This work was supported by the projects EU-ICT IntellIoT (grant agreement No. 957218) and Infotech-R2D2.
} 
}

\maketitle

\begin{abstract}

This work presents a harmonic design of \gls{agv} control, edge intelligence, and human input to enable autonomous transportation in industrial environments.
The \gls{agv} has the capability to navigate between a source and destinations and pick/place objects.
The human input implicitly provides preferences of the destination and exact drop point, which are derived from an \gls{ai} module at the network edge and shared with the \gls{agv} over a wireless network.
The demonstration indicates that the proposed integrated design of hardware, software, and \gls{ai} design achieve a \gls{trl} of range 4-5.

\end{abstract}

\glsresetall
\section{Introduction}

The rapid growth of customer demands and increasing costs of resources, labor, and energy have driven industries to seek new technologies that improve productivity and efficiency.
In particular, modern logistics is one of the key players to adopt autonomous automation technologies including human-in-the-loop, in which, human operators can provide the preferences required for automation~\cite{Tang.21}.
\Glspl{agv} are one of the pivotal components to enable "smart logistics" in manufacturing plants~\cite{Li.28}.
Realizing full/semi-autonomy with the fusion of control systems, communication networks, and computation servers at the edge over repetitive tasks calls for \gls{ai}-based solutions~\cite{Shaw.19}.
%


With the increasing interests in utilizing \glspl{agv} in industrial applications, developing efficient and reliable \gls{ai}-driven solutions for navigation tasks~\cite{Shaw.19} is of utmost importance.
Towards this, line following robots have received significant attention due to their ease of use and low complexity and robust operations~\cite{Bosnak.21}.
In this work, we demonstrate the codesign of an intelligent crawler robot, which uses its camera to sense the environment and, navigate and accurately deliver objects to their corresponding location with the aid of an edge \gls{ai} server to meet preferences set by a human operator.


\section{System Architecture}\label{sec:architecture}

\begin{figure}[!t]
\centering
\includegraphics[width=\linewidth]{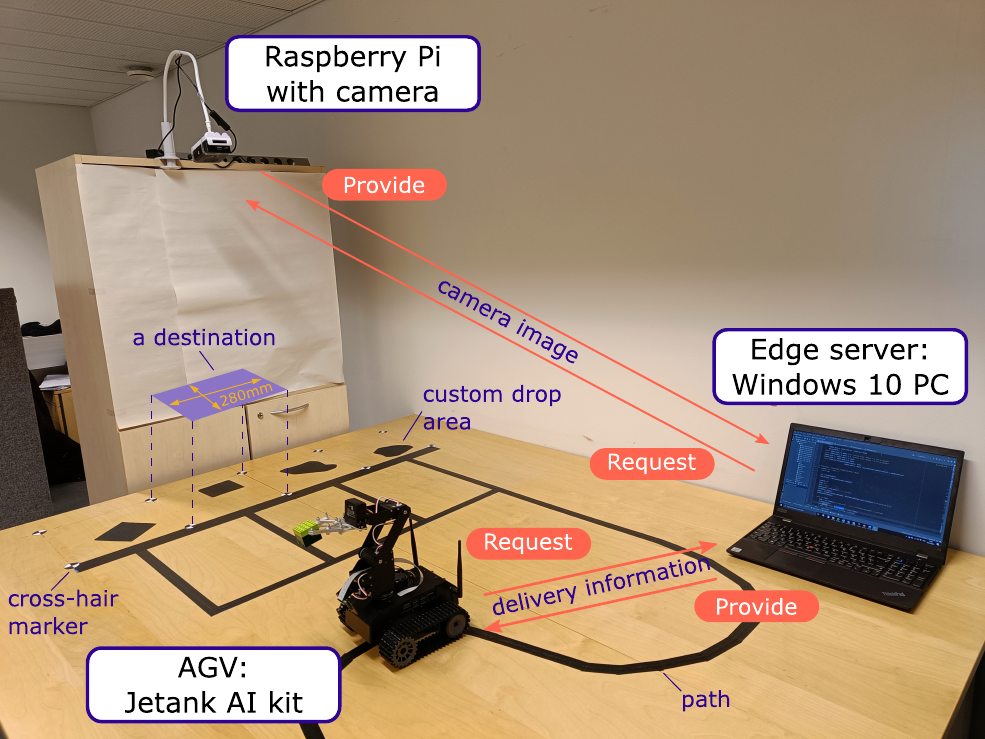}
\caption{The robotic platform highlighting the components and wireless connectivity.
}
\label{fig:sysmodel}
\end{figure}

\begin{figure*}[t]
\centering
\includegraphics[width=.8\linewidth]{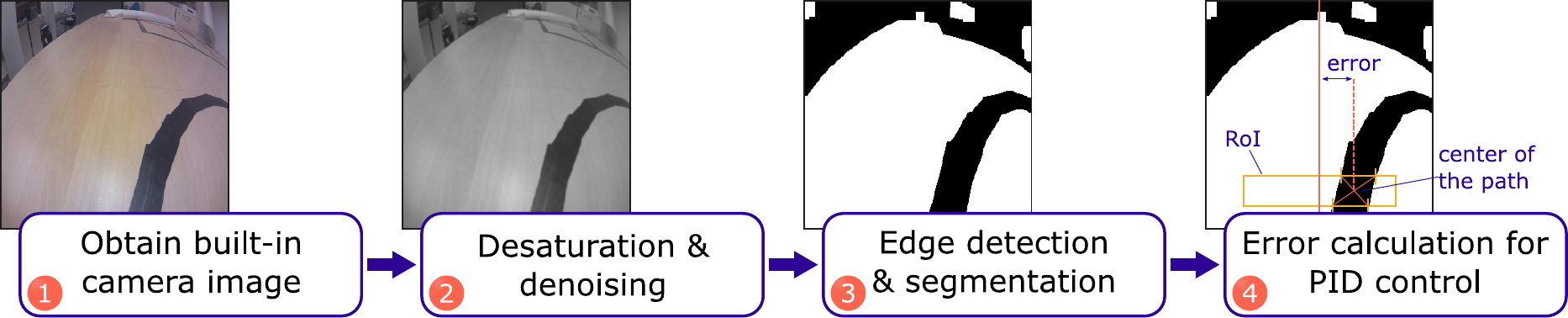}
\caption{The key steps of the intelligent navigation.}
\label{fig:nav_flow}
\end{figure*}

We consider the robotic platform illustrated in Fig.~\ref{fig:sysmodel} consisting of paths from a source point to multiple destinations, with an \gls{agv} that transports objects from source to destinations, a camera observing the destinations, and an edge server equipped with computing capabilities. 
The \gls{agv} uses an onboard camera to sense the environment and determine its control decisions, i.e., navigation among source and destinations.
The destination and precise drop location of the object, referred to as \emph{delivery information} hereinafter,
are determined by the edge server using camera images and shared with the \gls{agv} upon request.
The delivery information is derived based on custom drop areas defined by an external party (e.g., human operator).
Communication between the \gls{agv} and the edge server takes place through a REST API over a WiFi network.
The server provides delivery information in JSON format via its exposed endpoint, upon request by the \gls{agv} which takes place only once when each of the transport job is initiated. 
To derive delivery information, the edge server obtains a bird's eye view of the destinations by accessing the camera through its REST API over WiFi. 
After completing the delivery job successfully, the \gls{agv} returns to the source and repeats the procedure.

The paths between the source and the destinations are defined by black lines drawn on the surface to aid \gls{agv}'s navigation.
Here, a single path starting from the source, branches out to four paths leading to four adjacent destinations. 
Each destination is a square of width 280\,mm and the corners of the area are marked by cross-hair markers.
Hence, four destinations are defined by ten cross-hair markers as shown in Fig. \ref{fig:sysmodel} with the furthest marker at the left defined as the origin of the 2D coordinates.
The custom drop areas are located inside each of the destination, which are possibly irregular shapes.
The precise  drop location is defined as the center point of the largest circle placed inside the custom drop area.
It is worth highlighting that the drop point coordinates need to be computed with respect to the actual coordinates using a camera image having measurements in \gls{px}.
The placement of cross-hair markers is predefined and this knowledge is utilized to translate the \gls{px} distances to the actual measurements during image processing at the edge server. 


\section{System Implementation}

The demo setup is composed of several hardware components and two key software solutions providing the intelligence (i) for navigation and (ii) object placement. 
The details of these components and the implementation  are discussed next.

\subsection{Hardware}\label{sec:hardware}


Fig.~\ref{fig:sysmodel} shows all the hardware components used in the system.
\gls{agv} is a an off-the-shelf mobile robot known as "Jetank \gls{ai} kit” which is powered by Nvidia Jetson nano developer module with 16GB eMMC and 4GB RAM ~\cite{jetank}. 
The robot is equipped with 4-DoF mechanical arm and a wide-angle camera with $160^\circ$ field-of-view. 
The camera consists of a Raspberry Pi V2 camera module and connects with a Raspberry Pi 4 Model B computer, which hosts a web-server that serves images of the storage area upon the requests from the edge server.
A powerful multi-purpose 64-bit Windows 10 computer acts as the edge server and it hosts the \gls{ai} service and share the delivery information with the \gls{agv}. 
%


\begin{figure*}[!t]
\centering
\includegraphics[width=.8\linewidth]{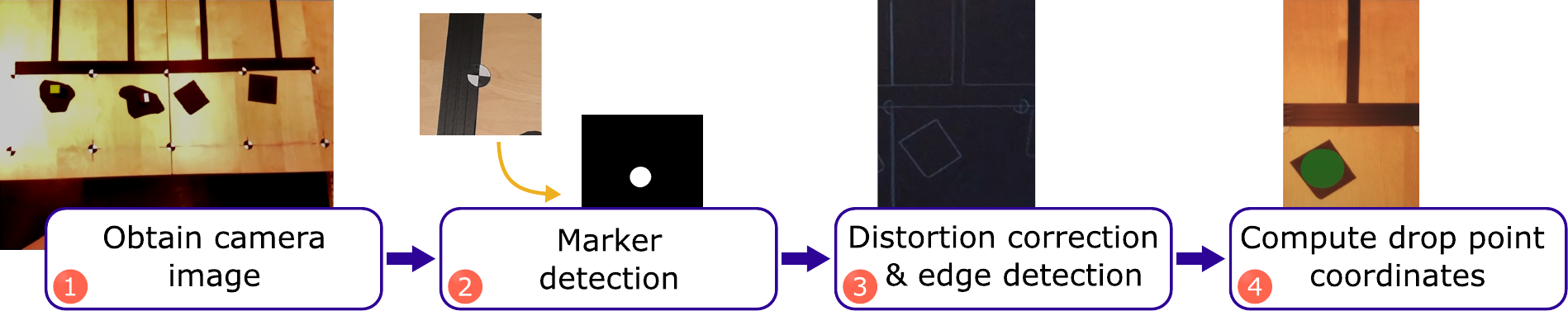}
\caption{The key steps of the edge \gls{ai} service.}
\label{fig:ai_flow}
\end{figure*}

\subsection{Intelligence for The Navigation}
\label{sec:ai_robot}

A vision-based line following system is implemented on the \gls{agv} to successfully navigate around the platform by using the on-board camera. 
The camera and the Jetson nano board support processing images with resolution $300\times300$ \gls{px} up to 30 frames per second.
During the navigation, the following three steps are repeated: 
i) acquiring camera images, 
ii) image pre-processing and line detection, and 
iii) actuating control decisions of the \gls{agv}
as illustrated in Fig. \ref{fig:nav_flow}.

The initialization is to set both the robot arm and the camera orientation to a position providing an obstruction free view of the path.
During navigation, camera images are obtained repeatedly, and converted to gray scale images.
Using a $3\times3$ Gaussian kernel, Gaussian blurring is performed to reduce the noise in the image (see step 2 in Fig. \ref{fig:nav_flow}).
Next, two morphological operators, erosion and dilation, are applied on the image to remove miniature inconsistencies on the detected path caused by to the glare of the surrounding light sources while preserving the structure and the shape of the path.
%
Then, the image is segmented by using OTSU method~\cite{Otsu.79}, which results in a binary image as shown under step 3 in Fig. \ref{fig:nav_flow}.
Therein, the path (dark surface) and the horizon (low lighting) of the image appear in black color while the rest appears in white. 
To neglect the undesired black areas corresponding to the horizon, a rectangular window that is likely to contain the path has been defined as the \emph{\gls{roi}} (see step 4 of Fig. \ref{fig:nav_flow}).
By assigning weights ones and zeros to black and white \gls{px}, respectively, the centroid of the \gls{roi} is calculated and referred to as the \emph{center of the path}.
The displacement of the center of the path from the vertical symmetrical axis of the image is denoted as the \emph{error}.
This error value is used in a \gls{pid} controller to determine the control commands (i.e., angular velocities of left and right wheels) ensuring a smooth navigation \cite{pid_controller}.
The coefficients of the proportional, derivative, and integral are 1, 1, and 0, respectively, which are obtained during the tuning phase of the \gls{agv}.

%
%

At junctions, the \gls{agv} switches to an alternative control procedure that relies on the type of the junction and the knowledge on the direction of moving and the delivery information. 
The junction is detected and its type is determined by analyzing the boundaries of the region of interest.
Next, based on the current goal, predefined sequences of control commands are issued to turn the \gls{agv} by $90^\circ$ or $180^\circ$.
%
After turning, the \gls{agv} hands over the control to the \gls{pid} algorithm. 
%
%
%
The \gls{agv} identifies the terminal points, source and destinations, by the absence of path segments in the \gls{roi}.
As the \gls{agv} reaches to source or destination, it activates pick/drop procedures accordingly. 
Here, the robot arm is programmed to move to a given 2D coordinate along its moving plane using inverse kinematics and grab or release the object.

\subsection{Intelligence for The Object Placement}
\label{sec:ai_edge}

The role of the edge server is to provide the intelligence in terms of computing the delivery information, which includes the destination and the drop location.
The destination is one of the four square areas defined by cross-hair markers. 
A human operator defines a custom drop area by placing a dark and possibly irregular flat surface made out of paper. 
The drop point is the center of the largest circle (not necessarily to be unique) that can be placed inside the user-defined area. 
Given the camera image, derivation of the delivery information requires several steps including cross-hair marker detection,
isolate the destination and camera distortion correction, identify the custom area within the destination,
and
compute drop coordinates in the \gls{px} domain and translate it to the actual coordinates.  
The flow of the edge server operation is illustrated in Fig.~\ref{fig:ai_flow}.

A pretrained \gls{ml} model is used for the cross-hair marker detection.
For the supervised training, $256\times256$ \gls{px} images extracted from the camera is used as the inputs while the labels are images with the same dimensions having white areas corresponding to the cross-hair markers and black areas reflecting everything else. 
To avoid handcrafting a dataset, a data augmentation procedure is adopted. 
First, using a single camera image (see step 1 in Fig.~\ref{fig:ai_flow} as an example) as the master sample, a master label is generated by filtering out the background (anything except cross-hair markers).
Then, different sizes of rectangles with different orientation are extracted randomly from the master sample and correspondingly from the master label. 
The extracted samples and labels are then resized to $256\times256$ \gls{px} images to generate the training and testing datasets of sizes 1024 and 128, respectively.
Next, a \gls{ml} model based on the U-NET architecture~\cite{ronneberger2015u} with the input and output dimensions of $256\times256$ is trained and tested with the aforementioned dataset. 
During inference, the camera image is partitioned into 10 segments, resized, and fed into the cross-hair marker detection model.
Once the cross-hair markers are detected (see step 2 in Fig.~\ref{fig:ai_flow}), the four destinations can be isolated. 
For the predefined destination, using the prior knowledge of markers to define a square, the detected markers are used to remove camera distortions.
The main benefit of the marker detection capability is that the service provided by the edge server is robust against camera reposition and variations in lighting conditions up to a certain limit.

The isolated destination is extracted as a separate image segment and the edge detection procedure based on OTSU method explained in Sec.~\ref{sec:ai_robot} is used to identify the edges of the custom drop area as illustrated under step 3 in Fig.~\ref{fig:ai_flow}. 
To determine the drop point, which is the center of the largest circle that can be placed inside the drop area (e.g., step 4 in Fig.~\ref{fig:ai_flow}), the edges are first approximated in to a polygon.
Then, the desired point is the furthest point from all the edges that lies inside the polygon.
This geometric solution is provided as a built-in function named \texttt{polylabel()} in a python package named poly-label~\cite{poly_label}.
Since the drop point coordinates are returned in \gls{px} with respect to the extracted image segment of the destination, they are translated to the actual physical coordinates based on prior knowledge of the setup dimensions and then shared with the \gls{agv}.

\subsection{Demo, Resources, and Future Developments}

The software related to this demo is available at github \url{https://github.com/ICONgroupCWC/Demo.Percom23}.
With a similar platform developed as per specifications provided in Sec.~\ref{sec:architecture} and the use of the hardware specified under Sec.~\ref{sec:hardware} along the aforementioned software, this demo can be reproduced.
This demo in action can be seen from \url{https://youtu.be/DhCSCCZbuHo}.
%

The lighting conditions (over/under-exposure and harsh shadows) directly impact the performance of the \gls{agv}.
To improve the overall performance, brightness and, exposure corrections methods and deep learning models for denoising and filtering will be investigated in future.
%

\bibliographystyle{IEEEtran}
\bibliography{reference}

\end{document}